\documentclass[letterpaper]{article} 
\usepackage{aaai25}  
\usepackage{times}  
\usepackage{helvet}  
\usepackage{courier}  
\usepackage[hyphens]{url}  
\usepackage{graphicx} 
\usepackage{natbib}  
\usepackage{caption} 
\usepackage{algorithm}
\usepackage{algorithmic}

\usepackage{amsmath}
\usepackage{booktabs}
\usepackage{pifont}
\usepackage{amssymb}
\usepackage{multirow}
\usepackage{makecell}
\usepackage{enumerate}
\usepackage{threeparttable}
\usepackage{newfloat}
\usepackage{listings}
\DeclareCaptionStyle{ruled}{labelfont=normalfont,labelsep=colon,strut=off} 
\floatstyle{ruled}
\newfloat{listing}{tb}{lst}{}
\floatname{listing}{Listing}
\title{PVTree: Realistic and Controllable Palm Vein Generation for Recognition Tasks}
\author{
    Sheng Shang\textsuperscript{\rm 1}\footnotemark[1],
    Chenglong Zhao\textsuperscript{\rm 2}\footnotemark[1],
    Ruixin Zhang\textsuperscript{\rm 2},
    Jianlong Jin\textsuperscript{\rm 1},
    Jingyun Zhang\textsuperscript{\rm 3}, \\
    Rizen Guo\textsuperscript{\rm 3}, 
    Shouhong Ding\textsuperscript{\rm 2}\footnotemark[2],
    Yunsheng Wu\textsuperscript{\rm 2},
    Yang Zhao\textsuperscript{\rm 1},
    Wei Jia\textsuperscript{\rm 1}\footnotemark[2] 
}
\affiliations{
    \textsuperscript{\rm 1} School of Computer Science and Information Engineering, Hefei University of Technology, China \\
    \textsuperscript{\rm 2} Tencent Youtu Lab \\
    \textsuperscript{\rm 3} Tencent WeChat Pay Lab33 \\
    shengshang@mail.hfut.edu.cn,
    \{lornezhao, ruixinzhang\}@tencent.com,\\
    jianlong@mail.hfut.edu.cn, 
    \{naskyzhang,
    rizenguo,
    ericshding, simonwu\}@tencent.com, \\
    \{yzhao,
    jiawei\}@hfut.edu.cn

}

\usepackage{bibentry}

\begin{document}

\maketitle

\renewcommand{\thefootnote}{\fnsymbol{footnote}} 
\footnotetext[1]{Equal contribution.} 
\footnotetext[2]{Corresponding authors.} 

\begin{abstract}
Palm vein recognition is an emerging biometric technology that offers enhanced security and privacy. However, acquiring sufficient palm vein data for training deep learning-based recognition models is challenging due to the high costs of data collection and privacy protection constraints. This has led to a growing interest in generating pseudo-palm vein data using generative models. Existing methods, however, often produce unrealistic palm vein patterns or struggle with controlling identity and style attributes.
To address these issues, we propose a novel palm vein generation framework named PVTree. First, the palm vein identity is defined by a complex and authentic 3D palm vascular tree, created using an improved Constrained Constructive Optimization (CCO) algorithm. Second, palm vein patterns of the same identity are generated by projecting the same 3D vascular tree into 2D images from different views and converting them into realistic images using a generative model. As a result, PVTree satisfies the need for both identity consistency and intra-class diversity.
Extensive experiments conducted on several publicly available datasets demonstrate that our proposed palm vein generation method surpasses existing methods and achieves a higher $TAR@FAR=1e-4$ under the 1:1 Open-set protocol. 
To the best of our knowledge, this is the first time that the performance of a recognition model trained on synthetic palm vein data exceeds that of the recognition model trained on real data, which indicates that palm vein image generation research has a promising future.


\end{abstract}

\begin{links}
\link{Code}{https://github.com/Sunniva-Shang/PVTree-palm-vein-generation}
\link{Dataset}{https://github.com/Sunniva-Shang/Palm-Vein-Dataset-HFUT}
\end{links}

%

\section{Introduction}

Palm vein recognition has become a very promising biometrics technology. Unlike some external biometrics such as face, fingerprints, and iris, palm veins are located under the skin. Therefore, palm vein images are more difficult to collect and are less susceptible to unauthorized capture by others. This means that palm vein recognition technology has its own unique advantages in protecting the user's privacy and providing a higher level of security.
At the same time, the difficulty of collection and the limitation of protecting user's privacy have led to a lack of large-scale public datasets in the field of palm vein recognition. However, training a deep learning-based palm vein recognition model requires a large-scale palm vein dataset.

\begin{figure}[t]
   \centering
   \center{\includegraphics[width=1\linewidth]  {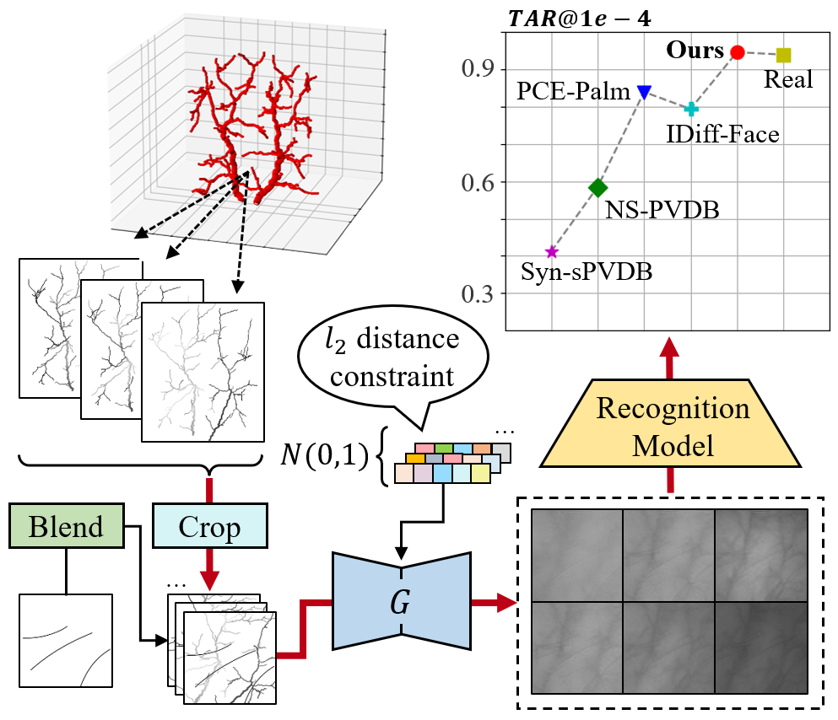}}
   \caption{The overall process of our method.}
   \label{fig:abs}
 
\end{figure}

Several methods have been proposed to generate pseudo-palm vein data\cite{salazar2021generating, salazar2021automatic, ou2022gan, salazar2024palm}. However, these methods present two obvious drawbacks:
1) The palm vein images generated by existing methods are not realistic enough. Some methods\cite{salazar2021automatic, salazar2021generating, ou2022gan} use palm vein pattern modeling methods or directly employ Generative Adversarial Networks(GANs) to generate palm vein images. However, 
most existing palm vein modeling methods fail to accurately capture the characteristics of real palm veins. Additionally, the training of GANs is frequently constrained by the small size of the available datasets, leading to the generation of less realistic palm vein images.
2) The palm vein images generated by existing methods lack intra-class variation. Some methods\cite{salazar2024palm} take into account the characteristics of palm veins for vein modeling and can generate visually realistic vein images. However, they lack the ability to control the degree of intra-class variation, which restricts the effectiveness of synthetic data in enhancing the performance of downstream recognition models.

In order to generate more realistic images of palm veins, we propose a novel and effective strategy for modeling the structure of palm veins, which is a 3D palm vascular tree generation framework named PVTree. It is designed to simulate palm vascular patterns better. Specifically, we introduce a common vascular tree modeling algorithm, Constrained Constructive Optimization (CCO) \cite{schreiner1993computer}, which combines the physiological and geometric characteristics of blood vessels to generate realistic vascular patterns. However, as a random generation method, CCO cannot meet the requirements for generating vascular networks with specific distributions. Therefore, we improve it to generate palm vascular networks.
We establish a vascular tree trunk that conforms to the characteristics of palm vascular patterns by combining palm anatomy and observations from a large amount of real data. This trunk can guide the CCO algorithm to generate vascular branches, thereby obtaining a vascular tree that satisfies specific rules. Subsequently, the generated 3D palm vascular tree is projected onto a 2D plane to acquire palm vein patterns as a new identity.
The proposed PVTree model effectively simulates the complex patterns of palm blood vessels and supports the generation of multiple types of vascular trees by altering the trunks based on empirical distribution.

In order to increase the intra-class variation of the generated palm vein images, we design several components to generate diverse samples for the same identity. Firstly, we propose a multi-view projector, which obtains projected palm vein patterns from different perspectives and vein depths. Secondly, we introduce multiple image enhancement techniques to simulate the effects of various palm postures. Subsequently, we adopt an image-to-image transfer model, PCE-Palm \cite{jin2024pce}, which represents the state-of-the-art in palmprint generation. This model not only supports the transfer of our palm vein patterns to the real image domain but also generates diverse samples. 
We further improve the method of generating diverse samples to achieve control over the degree of diversity.
A synthetic dataset containing samples of diverse styles can effectively improve the generalization ability of recognition models and enhance the potential for replacing real data.

The overall framework of our approach is illustrated in Figure \ref{fig:abs}. It effectively addresses the current challenges in the field of palm vein generation while also mitigating the issues of insufficient publicly available datasets and high data collection costs. Additionally, it fosters the advancement of using palm veins for biometric recognition. In summary, our contributions are as follows:
\begin{itemize}
\item We propose a novel framework for generating palm vein images that effectively models realistic vascular patterns and generates diverse samples for the same identity.
\item We propose a palm vascular tree modeling method named PVTree. To the best of our knowledge, this is the first method for modeling palm vein structures based on 3D modeling in the field of palm vein image generation. It effectively simulates the physiological characteristics and distribution patterns of palm blood vessels.
\item We collect a dataset with distinct palm veins, named HFUT, for palm vein research. 
A detailed description is provided in the dataset link.
\item We conduct extensive recognition experiments on several public palm vein datasets and synthetic datasets. The experimental results demonstrate that our method significantly outperforms existing methods.
Furthermore, it is the first time that the performance of the recognition model trained on our synthetic palm vein data exceeds that of the recognition model trained on real data.
\end{itemize}

\section{Related Work}
\subsection{Palm Vein Recognition Methods}
Palm vein recognition methods can be divided into two categories. The first category is the traditional methods\cite{wirayuda2015palm, wang2016single, rahul2015novel}, which usually use hand-crafted feature extractors to extract features such as structure, orientation, texture, etc. of the palm veins, which are then fed into a classifier for classification.
The second category is based on deep learning\cite{pan2019multi, wu2021outside, chen2021explainable, htet2023contactless}, primarily leveraging the powerful feature extraction capabilities of deep models to extract palm vein features and perform recognition. Compared to traditional methods, deep learning-based recognition methods typically yield higher recognition performance and have been widely employed in various authentication scenarios.
However, these methods require a large-scale size of training dataset.


\begin{figure*}[ht]
\centering
\includegraphics[width=0.98\linewidth] {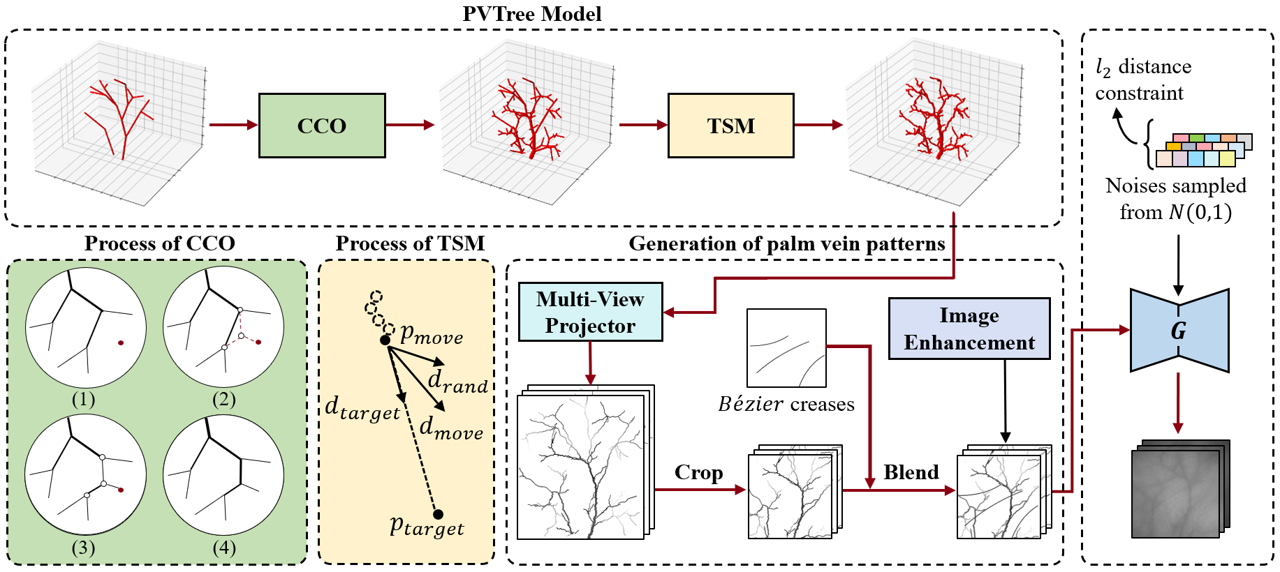}
\caption{Pipeline of the proposed method.
Firstly, PVTree Model is used to generate a palm vascular tree, which starts from the trunk representing vein types that we have summarized, and an authentic 3D palm vascular tree is obtained through the Constrained Constructive Optimization(CCO) and Trajectory Simulation Mechanism(TSM). 
Then Multi-views project, crop, and blend it with the B\'ezier creases\cite{zhao2022bezierpalm} representing palmprint to obtain the palm vein patterns representing identity. Finally, by improving the noise sampling mechanism of PCE-Palm, diverse palm vein images are generated.}
\label{fig:allpipline}
\end{figure*}

\subsection{Palm Vein Generation Methods}
GAN is an off-the-shelf generator, employed to generate palm vein data from latent space \cite{salazar2021generating}. In the case of insufficient training data, it is difficult to directly train GAN to learn the true distribution of the palm vein data. 
Some methods \cite{salazar2021automatic, ou2022gan} utilize the agent-based growth algorithm, such as Physarum \cite{baumgarten2010plasmodial} and Random Block Composition (RBC)\cite{ou2022gan}, to simulate vascular network patterns. However, these methods often exhibit a big gap between generated and real data. Another approach \cite{salazar2024palm} incorporates the physiological characteristics of blood vessels to generate visually realistic palm vein images. Diffusion model \cite{ho2020denoising} has increasingly showcased remarkable generative potential and has been successfully implemented in the field of biometric generation \cite{boutros2023idiff}.
However, these methods lack in-class control over the generation of samples.
Hence, there is an absence of a generation technique for palm veins that effectively combines both realism and abundant intra-class variations.

\subsection{Image-to-Image Translation Models}
Image-to-image translation models are designed to transform images from one domain to another. Paired Image-to-Image Translation \cite{isola2017image} is a conditional GAN that learns to map input images to output images using paired training data. CycleGAN \cite{zhu2017unpaired} is an unsupervised model that utilizes unpaired training data and introduces cycle consistency loss to ensure a consistent translation from one domain to another and back to the original domain.
ControlNet \cite{zhang2023adding} is an innovative diffusion model framework that permits the input of a conditioning image, which can then be employed to regulate the image generation process. These models have been used for tasks such as colorization, style transfer, and domain adaptation.
PCE-Palm \cite{jin2024pce}is proposed to generate palm print datasets that not only enable paired domain transfer but also support diverse variations under the same input. Consequently, PCE-Palm is more suitable for our palm vein generation compared to other methods.

\section{Methods}
In this section, we first discuss the anatomy of the palm vasculature and summarize the distribution features of vascular patterns. Based on these characteristics, we propose a 3D palm vascular tree modeling method named PVTree. Subsequently, palm vein patterns are generated by rendering the 3D palm vascular tree from multiple views. Finally, we employ an image-to-image transfer model to convert our patterns into palm vein images. The entire pipeline is illustrated in Figure \ref{fig:allpipline}.

\subsection{Palm Vascular Anatomy}

\begin{figure}[h]
   \centering
   \center{\includegraphics[width=0.99\linewidth] {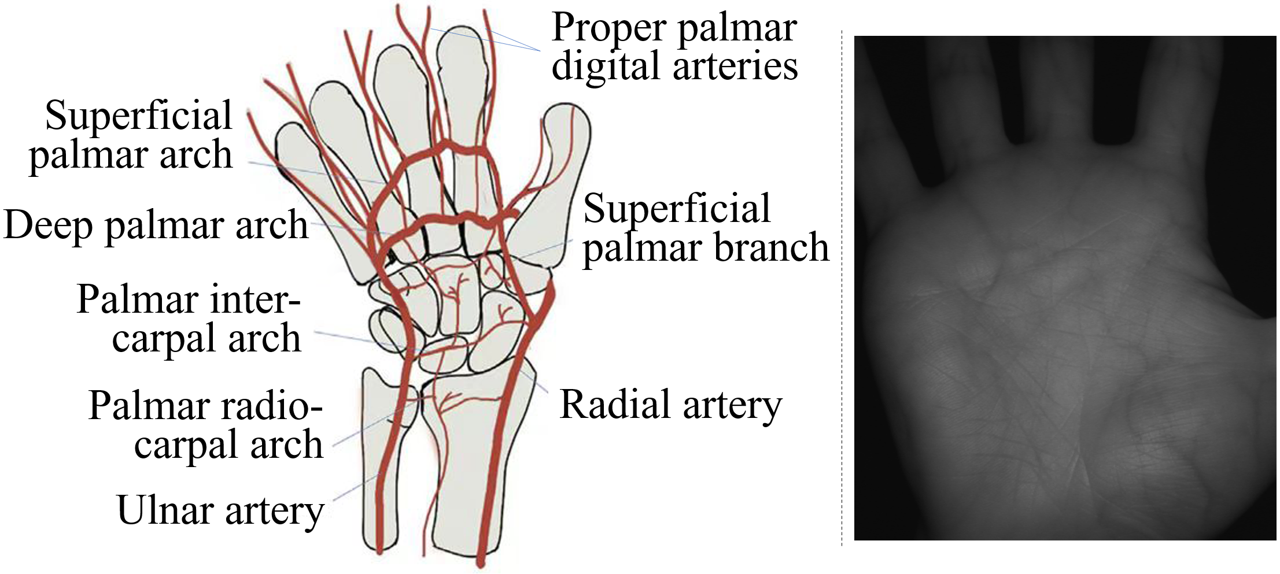}}
   \caption{Anatomical structure of palm blood vessels and their imaging under near-infrared light.
   \textbf{Left:} Vascular anatomical structure \cite{tan2020vascular};
   \textbf{Right:} A palm vein image from our dataset HFUT.
   }\label{fig:jiepou}
\end{figure}

Figure \ref{fig:jiepou} illustrates the anatomical structure of palm vasculature and a palm vein image under near-infrared light.  By combining these two elements, we highlight some notable features of palm vascular patterns:
\begin{enumerate}[1)]
\item The blood vessels of the palm are primarily supplied by two main arteries: the ulnar artery and the radial artery;

\item These two arteries interconnect, forming superficial and deep palmar arches, which create a complex vascular network in the palm area;

\item From these two palmar arches, numerous small blood vessels branch out to supply other parts of the fingers and palm;

\item Each finger has a pair of arteries that supply blood to both sides of the finger and the fingertips.

\end{enumerate}

\begin{figure}[h]
   \centering
   \center{\includegraphics[width=0.98\linewidth] {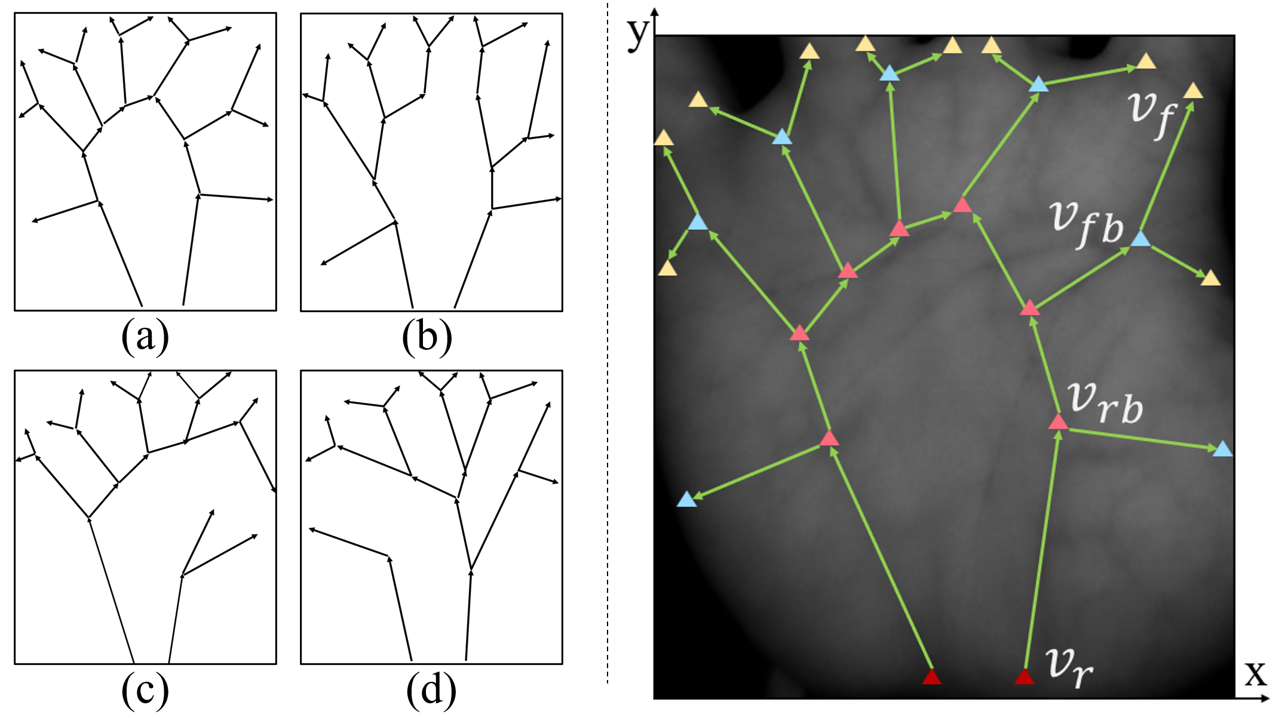}}
   \caption{Classification and construction of palm vascular tree trunks. 
   \textbf{Left:} (a),(b),(c),(d) are the classification of vascular trunks based on observations of a large amount of real data; 
   \textbf{Right:} Construction of type (a) trunk.
   The palm vein image comes from our dataset HFUT.}\label{fig:trunk}
\end{figure}

Based on the previous classification of palm vein patterns\cite{ottone2010analysis}, we categorize the patterns into four types.
As shown in Figure \ref{fig:trunk}, (a) represents the type where two arteries connect to form a palm arch that jointly supplies blood to other areas. Types (b), (c), and (d) lack a palm arch. Specifically, (b) represents the type where both arteries supply blood together, while (c) and (d) represent types where a single artery supplies blood. These classifications assist us in accurately modeling palm vein patterns.

\subsection{Proposed PVTree Model}
To better simulate palm vein patterns, we propose to construct a 3D palm vascular tree within the 3D space $\Omega$  defined as a rectangular prism,
\begin{equation}
    \Omega = (0, W) \times (40-D/2, 40+D/2) \times (0, H),    
\end{equation}
where, $W=70$, $D=14$, and $H=80$ represent the width, thickness, and height of the palm area, respectively.  
Based on the aforementioned distribution characteristics of palm vasculature, we adopt a two-stage method: first generating the trunks of the palm vascular tree, and then using the CCO algorithm to generate the vascular branches.

\textbf{Generation of the vascular tree's trunks} 
To ensure the authenticity of pattern modeling, the trunks are derived from real images. The right side of Figure \ref{fig:trunk} illustrates the construction of a type (a) trunk. We select the corresponding type of palm vein image from real datasets and crop it to match the palm area corresponding to $\Omega$, as indicated by the red box in Figure \ref{fig:trunk}.
Subsequently, we classify the key points of the palm vasculature into four types based on their distribution characteristics: $v_r$, $v_{rb}$, $v_{fb}$, and $v_{f}$. Among these, $v_{r}$ is the inflow point of the ulnar artery and radial artery, which form the palmar arch and extend to other areas. $v_{rb}$ is the bifurcation point where the palmar arch extends outward. $v_{fb}$ is the inflow point of the finger artery, and $v_f$ starts from $v_{fb}$ and flows into the fingers. 
These key points ensure the inflow and outflow directions of each vascular segment, as indicated by the arrows in Figure \ref{fig:trunk}.

After identifying the position of key points, we introduce some perturbation noise to these points. For each identity, we sample the positions of the key points within the perturbation range to generate different trunks.
Subsequently, we transform each key point $v(x_{2d}, y_{2d})$ into the $\Omega$ space to obtain a three-dimensional representation $v(x_{3d}, y_{3d}, z_{3d})$, defined as follows:
\begin{equation} 
    \begin{cases}
        x_{3d} = x_{2d}; \\
        y_{3d} = Uniform(40-D/2, 40+D/2);  \\
        z_{3d} = y_{2d}.
    \end{cases}
\end{equation}

In addition, for the radius optimization of each vascular segment, we establish its relationship with the outflow of the segment as follows:
\begin{equation}
  r_i = r_{0} + n_i * ratioE,
\end{equation}
where $r_{0}$ is the initial radius, $n_i$ is the number of endpoints that flow out of the vascular segment $s_i$, and $ratioE$ is its weight. Thus, we complete the construction of the trunk. 
The construction methods for types (b), (c), and (d) are the same.

\textbf{Generation of vascular tree's branches} 
Based on the previously mentioned vascular tree trunks, we iteratively add $N=70$ points within $\Omega$ to connect with the trunk and form vascular tree branches. Here, we introduce the CCO algorithm, which facilitates finding the optimal position for a new bifurcation point under the constraints of optimization objectives, by minimizing the total volume of the vascular tree.

\begin{figure}[h]
   \centering
   \center{\includegraphics[width=0.4\linewidth] {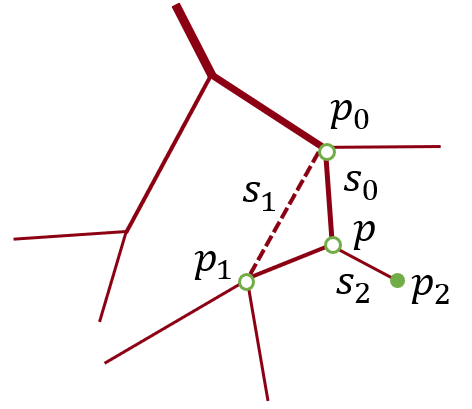}}
   \caption{The formation process of vascular tree’s branches.}\label{fig:kamyia}
\end{figure}

As shown in Figure \ref{fig:kamyia}, $p_{2}$ is a new point that will be connected to its nearest vein segment, $s_1$. The bifurcation point $p$ is located within the triangle formed by the two endpoints, $p_0$ and $p_1$, of $s_1$ and the new point $p_{2}$. Subsequently, we need to connect $p_{2}$ and $p$ to the existing vascular tree.
In this process, segment $s_1$ is removed and reconnected with $p$ and $p_1$, while the bifurcation point $p$ is connected to both $p_0$ and $p_{2}$, forming new segments $s_0$ and $s_2$. The radius of the new segments and the position of $p$ are determined according to the Kamiya optimization algorithm \cite{kamiya1972optimal}.

In addition, because each vascular segment is modeled as a cylinder in 3D space, its projection is a line segment. However, real palm vascular patterns are often composed of curves. To better simulate this, we propose a Trajectory Simulation Mechanism.
Let $T$ be a generated 3D palm vascular tree, $s$ be one of its vascular segments, and $p_1$ and $p_2$ be the inflow and outflow points of the segment, respectively. We consider the inflow point $p_1$ as a moving point $p_{move}$, moving towards the target direction $d_{target} = p_{move} - p_2$, with a step size $l_{step}$ for each movement. During each movement, we apply a random disturbance $d_{rand}$ to its main direction $d$. Thus, the position $P_i$ after the $i$-th movement is obtained as:
\begin{equation}
    P_{i} = P_{i-1} + l_{step} * (d_{target} + w_{rand} * d_{rand}),
\end{equation}
where $w_{rand}$ is the weight of the disturbance.
When the step $l_{step}$ is small enough, we obtain an approximate curved trajectory from $p_1$ to $p_2$.

\subsection{Intra-Class Variation} 

During the collection of palm vein datasets, images of the same identity are often influenced by factors such as acquisition viewpoint, vein depth, palm posture, collection environment, and palmprint. 
To simulate the influence of palmprints, we blend the B\'ezier crease \cite{zhao2022bezierpalm} with the projected vascular patterns. 
For the other influencing factors, we introduce several control components to simulate them.


\textbf{Disturbances in 3D projection}
The 3D palm vascular tree offers a natural advantage in addressing the effects of acquisition viewpoint and simulating vein depth. By rotating the 3D palm vascular tree around the $z$-axis, we obtain the projected images from different viewpoints. 
It is noted that larger rotation angles may cause significant changes in the patterns representing identity, which is not conducive to maintaining identity uniqueness. After multiple tests, we found that the optimal rotation range for the vascular tree is between -3 and 3 degrees.

In addition, for the y-axis value representing the depth of blood vessels, we establish its relationship with the projected grayscale value $G$ as,

\begin{equation}
    G = \frac{(y-y_{min}) * w_{random}}{y_{max}-y_{min}} \times 255,
\end{equation}
where $y_{max}$ and $y_{min}$ are the maximum and minimum depths of all vascular segments in the 3D palm vascular tree, and $w_{random}$ is a random perturbation that induces slight grayscale value changes in each projection, corresponding to the depth variations of the veins.

\textbf{Disturbances in palm vein patterns}
To simulate the influence of palm posture, we employ various image enhancement techniques. Specifically, we apply minor random scaling, rotation, distortion, and cropping to the palm vein patterns, generating variations that reflect different postures for the same identity.

\textbf{Constraint on the distance between noises}
For intra-class variations, we improve the noise sampling mechanism in PCE-Palm. Instead of the original random sampling from a normal distribution, we use noise sampling with $l_2$ distance constraints.
By adjusting the $l_2$ distance, we obtain samples with different degrees.

\subsection{Generation of Palm Vein Images}
Once the palm vein patterns are created, it is crucial to render them into realistic images. To do this, we utilize an image-to-image translation model, i.e., PCE-Palm. To better adapt it to our task, we improve its inference phase.

The original PCE-Palm model first transfers the patterns to the Palm Crease Energy (PCE) domain during the inference stage to narrow the domain gap, and then inputs them into the generator to obtain the final image. 
Here, we remove this process and directly use our generated palm vein patterns as the input of the generator. 
This modification is based on the observation that real palm vein images usually exhibit local blurring characteristics due to varying vein depths. So the palm vein patterns extracted by the PCE extraction module (PCEM) represent only the most prominent parts of the veins rather than the complete patterns.
A generative model trained with such non-depth lines or purely black lines implies that the dark line areas correspond to the prominent vein areas in the synthetic images. In contrast, the palm vein patterns generated by our PVTree model contain complete patterns with depth information.
As a result, when these patterns are directly used for inference, the generated image's prominent vein areas correspond to the regions with high vein depth information in the patterns. This effectively preserves the depth information of the veins, enabling a better simulation of the local blurring characteristics observed in real palm veins.

\section{Experiment}
\subsection{Experimental Setup}
\textbf{Datasets} 
We utilize four publicly available palm vein datasets: CASIA \cite{hao2008multispectral}, PolyU \cite{zhang2009online}, TongJi \cite{zhang2018palmprint}, PUT \cite{kabacinski2011vein}, and our own HFUT dataset, totaling 1,602 identities and 25,248 images. Detailed information about these public datasets is shown in Table \ref{tab:pubdata}.
Among them, CASIA is a multispectral dataset, and we use only the images captured at a wavelength of 940nm. The PUT dataset does not contain complete hand images, making it impossible to extract ROI images using the standard ROI extraction algorithm. Therefore, we use the PUT dataset solely for training the generative model, not the recognition model.
Following the open-set protocol, we divide each dataset into training and testing sets at a 1:1 ratio. For the generative model, all images from the PUT dataset are included in the training set, comprising 851 identities and 13,224 images. For the recognition model, we use the training sets of the other four public datasets, totaling 751 identities and 12,024 images.

\begin{table}[h]
\small
\centering
\renewcommand\arraystretch{1.5}
\begin{tabular}{lcccc}
\toprule[0.5mm]
Datasets & IDs & Samples & Images & Sessions interval \\
\hline
CASIA & 200 & 6 & 1200 & 30days \\
PUT & 100 & 12 & 1200 & 7days \\
HFUT & 202 & 24 & 4848 & 7days \\
PolyU & 500 & 12 & 6000 & 9days \\
TongJi & 600 & 20 & 12000 & 61days \\
\toprule[0.5mm]
\end{tabular}
\caption{Details of the datasets used in our experiments.}
\label{tab:pubdata}
\end{table}

\begin{figure}[ht]
   \centering
   \center{\includegraphics[width=0.95\linewidth] {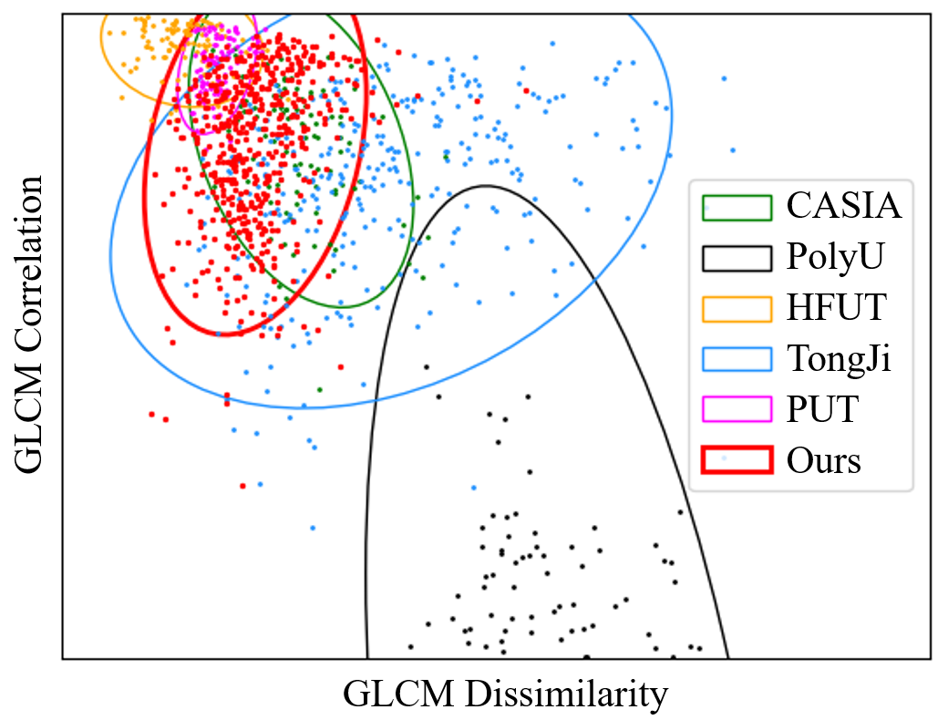}}
   \caption{The covariance confidence ellipse plot of two important GLCM features 
from our dataset and the real datasets.
}\label{fig:glcm}
\end{figure}

\textbf{Generation Model Training Setups} 
For the loss function weights and learning rate settings of the generative model, we follow the original PCE-Palm configuration\cite{jin2024pce}.
In addition, the model is trained for a total of 60 epochs, with a learning rate of $2 \times 10^{-4}$ for the first 30 epochs, linearly decaying to $1 \times 10^{-6}$ for the last 30 epochs. The resolution of the output image is 128 $\times$ 128, 
which is consistent with those of most images in the real datasets.

\begin{table}[ht]
\small
\centering
\renewcommand\arraystretch{1.5}
\begin{tabular}{lcccc}

\toprule[0.5mm]
Datasets & Syn-sPVDB & NS-PVDB & Ours \\
\hline
FID & 56.1685 & 50.3706 & \textbf{37.0820}\\
Wang17 &  0.4747 & 0.6065 & \textbf{0.6223} \\

\toprule[0.5mm]
\end{tabular}
\caption{Scores of quality assessment on real and synthetic datasets.}
\label{tab:quality}
\end{table}

\textbf{Recognition Model Training Setups} 
The network backbone of the recognition model is ResNet101 \cite{he2016deep}.
We use ArcFace \cite{deng2019arcface} with a scale factor of $s=64$ and a margin of $m=0.5$ to train for 20 epochs on both synthetic and real datasets. We employ a cosine learning rate schedule with one warmup epoch, setting the maximum and minimum learning rates to $1 \times 10^{-2}$ and $1 \times 10^{-6}$, respectively. All experiments are conducted on 8 GPUs with a batch size of 32.

\begin{table*}[ht]
\small
\centering
\renewcommand\arraystretch{1.5}
\begin{threeparttable}
\begin{tabular}{lcccccccccccc}
\toprule[0.5mm]
\multicolumn{3}{c}{Datasets} & \multicolumn{2}{c}{CASIA} & \multicolumn{2}{c}{PolyU} & \multicolumn{2}{c}{HFUT} & \multicolumn{2}{c}{TongJi} & \multicolumn{2}{c}{Average} \\
\hline
Name & IDs & Imgs & EER & \makecell{TAR@\\1e-4} & EER & \makecell{TAR@\\1e-4} & EER & \makecell{TAR@\\1e-4} & EER & \makecell{TAR@\\1e-4} & EER & \makecell{TAR@\\1e-4} \\
\hline 
Real data & 751 & 12024 & 0.0210 & 0.8988 & 0.0009 & 0.9962 & 0.0080 & 0.9352 & 0.0126 & 0.9217 & 0.0106 & 0.9380 \\
NS-PVDB & 4000 & 28000 & 0.0685 & 0.4669 & 0.0318 & 0.7333 & 0.0201 & 0.8130 & 0.1113 & 0.3241 & 0.0581 & 0.5843 \\
Syn-sPVDB & 4000 & 28000 & 0.2438 & 0.1461 & 0.0232 & 0.8424 & 0.0752 & 0.5387 & 0.2719 & 0.1199 & 0.1535& 0.4118 \\
PCE-Palm & 4000 & 28000 & 0.0454 & 0.7093 & 0.0070 & 0.9362 & 0.0095 & 0.8295 & 0.0183 & 0.8782 & 0.0201 & 0.8383 \\
IDiff-Face & 4000 & 28000 & 0.0398 & 0.6927 & 0.0042 & 0.9742 & 0.0188 & 0.8295 & 0.0460 & 0.6750 & 0.0272 & 0.7928 \\
\hline
\multirow{4}{*}{Ours} & 1000 & 7000 & \textbf{0.0188} & 0.8901 & 0.0008 & 0.9956 & 0.0111 & 0.9280 & 0.0158 & 0.9080 & 0.0116 & 0.9304 \\
\multirow{4}{*}{} & 2000 & 14000 & 0.0216 & 0.8821 & 0.0013 & 0.9921 & \textbf{0.0071} & \textbf{0.9421} & 0.0134 & 0.9199 & 0.0108 & 0.9341 \\
\multirow{4}{*}{} & 3000 & 21000 & 0.0210 & \textbf{0.9053} & 0.0021 & 0.9881 & 0.0087 & 0.9327 & 0.0129 & 0.9330 & 0.0112 & 0.9398 \\
\multirow{4}{*}{} & 4000 & 28000 & 0.0210 & 0.9038 & \textbf{0.0006} & \textbf{0.9969} & 0.0078 & 0.9346 & \textbf{0.0106} & \textbf{0.9437} & \textbf{0.0010} & \textbf{0.9448} \\

\hline
Mix\tnote{1} & 4751 & 40024 & \textbf{0.0159} & \textbf{0.9487} & \textbf{0.0004} & \textbf{0.9994} & \textbf{0.0014} & \textbf{0.9594} & \textbf{0.0062} & \textbf{0.9980} & \textbf{0.0060} & \textbf{0.9713} \\

\toprule[0.5mm]

\end{tabular}
\begin{tablenotes}
    \footnotesize
    \item[1] Mixing the real data with our 4000 IDs dataset.
    \end{tablenotes}
    \end{threeparttable}
    \label{tb:1}%
    
\caption{Recognition performance on real datasets and synthetic datasets under different settings.}
\label{tab:recog}
\end{table*}

\subsection{Assessment of Image Quality}

\begin{figure}[ht]
   \centering
   \center{\includegraphics[width=1\linewidth] {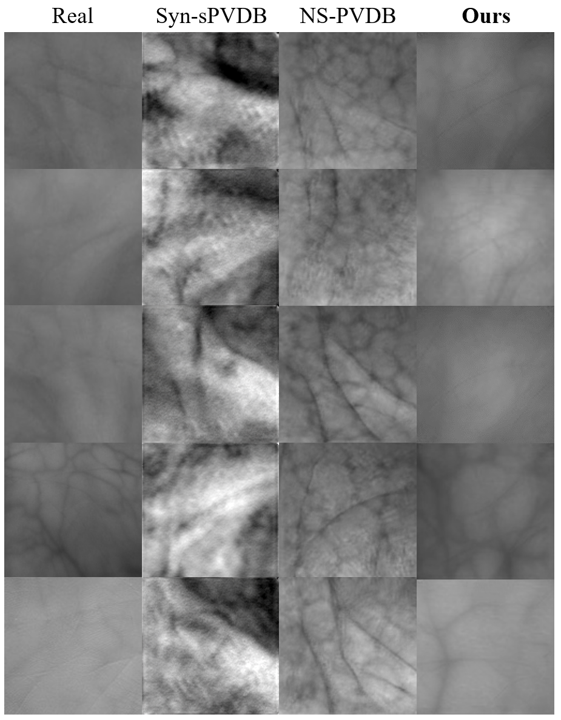}}
   \caption{Samples of our data with existing methods and real data.}\label{fig:comp}
\end{figure}

To validate the authenticity of the palm vein patterns modeled by PVTree, we use the Gray-Level Co-occurrence Matrix (GLCM) to compare the similarity between the generated patterns and the real data patterns. 
As shown in Figure \ref{fig:glcm}, a large overlap between the ellipses indicates a higher pattern similarity between the datasets. We observe that there are certain differences in the patterns of different real palm vein datasets, especially for PolyU, which uses a contact-based acquisition method that differs from others and makes the collected images more susceptible to interference from skin texture and device pressure.
The confidence ellipse of our synthetic dataset is located in the dense area of all ellipses and has significant overlap with the ellipses of real datasets. This demonstrates the similarity of the patterns between our dataset and the real datasets, further validating the authenticity and diversity of the vein pattern modeling method we proposed.

In addition, we assess image quality using both visual inspection and quantitative evaluation methods. Since some methods \cite{ou2022gan, salazar2024palm} have not publicly released their synthetic datasets and source code, we are unable to compare our data with theirs.
Figure \ref{fig:comp} shows samples of our data alongside Syn-sPVDB \cite{salazar2021generating}, NS-PVDB \cite{salazar2021automatic}, and real data. It can be observed that our data exhibits a higher similarity to the real data, effectively modeling the complex patterns and depth variations of veins. For quantitative assessment, we employ the Fréchet Inception Distance (FID) and the vein image-specific quality metric Wang17 \cite{wang2017quality} to evaluate the datasets.
The experimental results, as shown in Table \ref{tab:quality}, indicate that our data achieves the lowest FID score, underscoring its similarity to the real data. Furthermore, the comparison of vein image quality scores using Wang17 highlights the high quality of our data.

\begin{table}[ht]
\small
\centering
\renewcommand\arraystretch{1.5}
\begin{tabular}{p{5pt}p{3pt}p{3pt}cccccccc}
\toprule[0.5mm]
\multicolumn{3}{c}{Settings} & \multicolumn{5}{c}{TAR@1e-4}\\
N & R & A & CASIA & PolyU & HFUT & TongJi & Average \\

\hline
0.5 & \ding{55} & \ding{55} & 0.8497 & 0.9862 & 0.9284 & 0.9313 & 0.9239 \\
0.3 & \ding{55} & \ding{55} & 0.8106 & 0.9852 & 0.9249 & 0.9023 & 0.9057 \\
0.7 & \ding{55} & \ding{55} & 0.8040 & 0.9849 & 0.9070 & 0.8865 & 0.8956 \\
0.5 & \checkmark & \ding{55} & 0.8620 & 0.9850 & \textbf{0.9402} & 0.9321 & 0.9298 \\
0.5 & \checkmark & \checkmark & \textbf{0.9038} & \textbf{0.9969} & 0.9346 & \textbf{0.9437} & \textbf{0.9448} \\

\toprule[0.5mm]
\end{tabular}
\caption{Recognition performance on synthetic datasets under different intra-class control methods.}
\label{tab:ablation}
\end{table}

\subsection{Assessment of Recognition Performance}
To validate the improvement of our synthetic data for the recognition tasks, we compare it with real data and existing generation methods, including the original PCE-Palm and IDiff-Face \cite{boutros2023idiff}. 
Since some methods \cite{ou2022gan, salazar2024palm} are not publicly available, we compare our data with Syn-sPVDB and NS-PVDB among the existing palm vein generation methods.
Both Syn-sPVDB and NS-PVDB include 7 samples, so we ensure a fair comparison by maintaining the same number of samples for all datasets. 
We train the recognition model and evaluate it on the test set of each public dataset. All recognition models are trained and tested using the same set of hyperparameters. We use EER and TAR@1e-4 as evaluation metrics, and the experimental results are shown in Table \ref{tab:recog}. 

As shown in Table \ref{tab:recog}, the recognition models trained on datasets generated by existing methods still fall significantly short of the recognition performance achieved with real data. 
However, when the number of IDs in the dataset generated by our method increases to 3000, the performance of the model trained with our synthetic data exceeds the performance of the model trained with real data.
Moreover, when real data and synthetic data are mixed to train the recognition model, the recognition model achieves better recognition performance. This demonstrates the effectiveness of our proposed method.

\subsection{Ablation Study}
We compare the effects of different control components for intra-class variation on recognition performance, with the experimental setup remaining the same as before. All experiments are conducted on the synthetic dataset with 4000 IDs and 7 samples.
We present the ablation results of different control component settings in Table \ref{tab:ablation}. For convenience, we use N, R, and A to denote the $l_2$ distance between the noise input to generative models, disturbances in palm vein patterns, and disturbances in 3D projection, respectively.
We first compare the effects of different $l_2$ distances for noise and find that the optimal distance is 0.5. Additionally, it is evident that both the image enhancement and the multi-view projector further enhance recognition performance.

\section{Conclusion}
In this paper, we propose a two-stage palm vein generation method aimed at creating more realistic and diverse datasets. This pipeline simplifies the process of generating palm vein images by breaking down the complex task into two more manageable subproblems: vein structure generation and vein image translation. 
Furthermore, the proposed method enables the controlled and diverse generation of images, facilitating the generation of identified and varied palm vein images.
Experimental results reveal that the proposed method outperforms all existing palm vein generation methods and even surpasses the recognition model trained on real data. We also find that combining synthetic data with real data for recognition leads to significant improvements in recognition performance, demonstrating the effectiveness of our synthetic data. Our work reduces the dependence of deep learning-based palm vein recognition methods on large-scale real data and promotes the development of the field of palm vein recognition.

\section{Acknowledgments}
This work is partly supported by the National Science and Technology Major Project under Grant 2020AAA0107300 and the grants of the National Natural Science Foundation of China, Nos.62076086, 62476077, 62272142.

\bibliography{aaai25}

\end{document}